\documentclass{article}

\usepackage[preprint]{neurips_2022}

\usepackage{algorithm} 
\usepackage{algorithmicx}  
\usepackage{algpseudocode}  
\usepackage{bbold} 
\usepackage{amsmath} 
\usepackage{mathrsfs}
\usepackage{graphicx} 
\usepackage{subfig}
\usepackage{natbib}
\setcitestyle{numbers,square}
\usepackage{wrapfig}
\usepackage{xcolor}
\usepackage[utf8]{inputenc} 
\usepackage[T1]{fontenc}    
\usepackage{hyperref}       
\usepackage{url}            
\usepackage{booktabs}       
\usepackage{amsfonts}       
\usepackage{nicefrac}       
\usepackage{microtype}      
\usepackage{xcolor}         

\title{ADT-SSL: Adaptive Dual-Threshold for Semi-Supervised Learning}

\author{
  Zechen Liang\\
  School of Computer Science and Cyber Engineering\\
  Guangzhou University\\
  Guangzhou 510006, P. R. China \\
  \texttt{liangzechen@e.gzhu.edu.cn} \\
   \And
   Yuan-Gen Wang
   \\
  School of Computer Science and Cyber Engineering\\
  Guangzhou University\\
  Guangzhou 510006, P. R. China  \\
   \texttt{wangyg@gzhu.edu.cn} \\
  \AND
  Wei Lu \\
  School of Computer Science and Engineering\\ Guangdong Province Key Laboratory of Information Security Technology\\
  Ministry of Education Key Laboratory of Machine Intelligence and Advanced Computing \\
  Sun Yat-sen University\\
 Guangzhou 510006, P. R. China\\
 \texttt{luwei3@mail.sysu.edu.cn}\\
  \AND
  Xiaochun Cao \\
  School of Cyber Science and Technology \\
  Shenzhen Campus, Sun Yat-sen University \\
  Shenzhen 518107, P. R. China \\
  \texttt{caoxiaochun@mail.sysu.edu.cn} \\
}

\begin{document}

\maketitle

\begin{abstract}
 Semi-Supervised Learning (SSL) has advanced classification tasks by inputting both labeled and unlabeled data to train a model jointly. However, existing SSL methods only consider the unlabeled data whose predictions are beyond a fixed threshold (e.g., 0.95), ignoring the valuable information from those less than 0.95. We argue that these discarded data have a large proportion and are usually of hard samples, thereby benefiting the model training. This paper proposes an Adaptive Dual-Threshold method for Semi-Supervised Learning (ADT-SSL). Except for the fixed threshold, ADT extracts another class-adaptive threshold from the labeled data to take full advantage of the unlabeled data whose predictions are less than 0.95 but more than the extracted one. Accordingly, we engage CE and $L_2$ loss functions to learn from these two types of unlabeled data, respectively. For highly similar unlabeled data, we further design a novel similar loss to make the prediction of the model consistency. Extensive experiments are conducted on benchmark datasets, including CIFAR-10, CIFAR-100, and SVHN. Experimental results show that the proposed ADT-SSL achieves state-of-the-art classification accuracy. 
\end{abstract}

\section{INTRODUCTION}
Depending on a large amount of labeled data and supervised learning, current deep neural networks have achieved great success in computer vision tasks, such as image classification, object detection, and semantic segmentation. However, massive labeled image data are not always available in some scenarios, such as medical images, aerospace images, and images from new born fields. In addition, labeling large dataset requires a lot of manpower and financial resources. Semi-Supervised Learning (SSL) has been proposed by training a model with limited labeled data and a large number of unlabeled data, thus gaining great attention in the literature. 

The core problem of existing SSL methods is how to extract as much information as possible from unlabeled data and use it to train the model effectively. To this end, leading SSL methods first perform data augmentation, and then engage the consistency regularization to control the model to have consistent output for different augmented forms of the same data \cite{laine2016temporal, sajjadi2016regularization, sohn2020fixmatch, berthelot2019mixmatch, berthelot2019remixmatch, miyato2018virtual, tarvainen2017mean, xie2020unsupervised, chen2020simple, zhang2020wcp}. Besides, entropy minimization strategy is often used to make the model output have low entropy. To our best knowledge, all the existing SSL methods adopt a fixed threshold (e.g., 0.95) to select unlabeled data with high enough confidence. For example, MixMatch \cite{berthelot2019mixmatch} takes advantage of sharpen function to encourage the entropy minimization of the model output. On the basis of MixMatch, Remixmatch \cite{berthelot2019remixmatch} introduces distribution alignment and augmentation anchoring, which makes the prediction of the model for unlabeled data more precise and the learning of unlabeled data more effective. FixMatch \cite{sohn2020fixmatch} generates the one-hot labels for the unlabeled data whose predictions exceed the fixed threshold and applies distribution alignment to them. SimPLE \cite{hu2021simple} strengthens the utilization of unlabeled data by adding the learning of similar unlabeled data. Although a high fixed threshold allows the model to focus on these unlabeled data which have high confidence, it means that those unlabeled data which cannot reach the threshold will be discarded. We argue that it is unreasonable for all classes to use the same high threshold. This is because the model has different learning states for different classes at different learning stages. The recently proposed FlexMatch \cite{zhang2021flexmatch} considers different learning states for different classes. Unfortunately,  FlexMatch uses the same number of unlabeled data for all classes.

Motivated by the above limitations, we propose an Adaptive Dual-Threshold method for Semi-Supervised Learning (ADT-SSL). In addition to a fixed threshold, we extract another class-adaptive threshold from the labeled data to form an Adaptive Dual-Threshold (ADT) scheme. In fact, the unlabeled data whose predictions are less than the fixed threshold but greater than the extracted one account for a large proportion. We argue that these discarded unlabeled data are often of hard samples and enable to boost the model training. We set the class-adaptive threshold to be the smallest prediction of all the labeled data of one class that can be correctly classified. As done in most SSL methods, Cross Entropy (CE) loss is used for learning from the unlabeled data whose predictions are greater than the fixed threshold. In particular, we engage $L_2$-norm loss to learn from our newly mined unlabeled data. Furthermore, we propose a novel similar loss to learn these highly similar unlabeled data. Extensive experiments and ablation study show that our ADT-SSL surpasses the state-of-the-art approaches by a large margin. The novelties of this work are as follows:
\begin{itemize}
\item We develop a new concept of class-adaptive threshold. Combining the fixed threshold with the adaptive one, we propose an adaptive dual-threshold method (ADT), which can not only make the model selectively learn from unlabeled data but also balance the learning process of all classes.
\item We propose a novel similar loss to further exploit the information among the highly similar unlabeled data, thereby boosting the model training.
\item We construct an overall loss function by assigning four different losses to four different types of data, leading to state-of-the-art classification accuracy.
\end{itemize}
\begin{figure}[tp]
	\centering    
	\includegraphics[width=\textwidth]{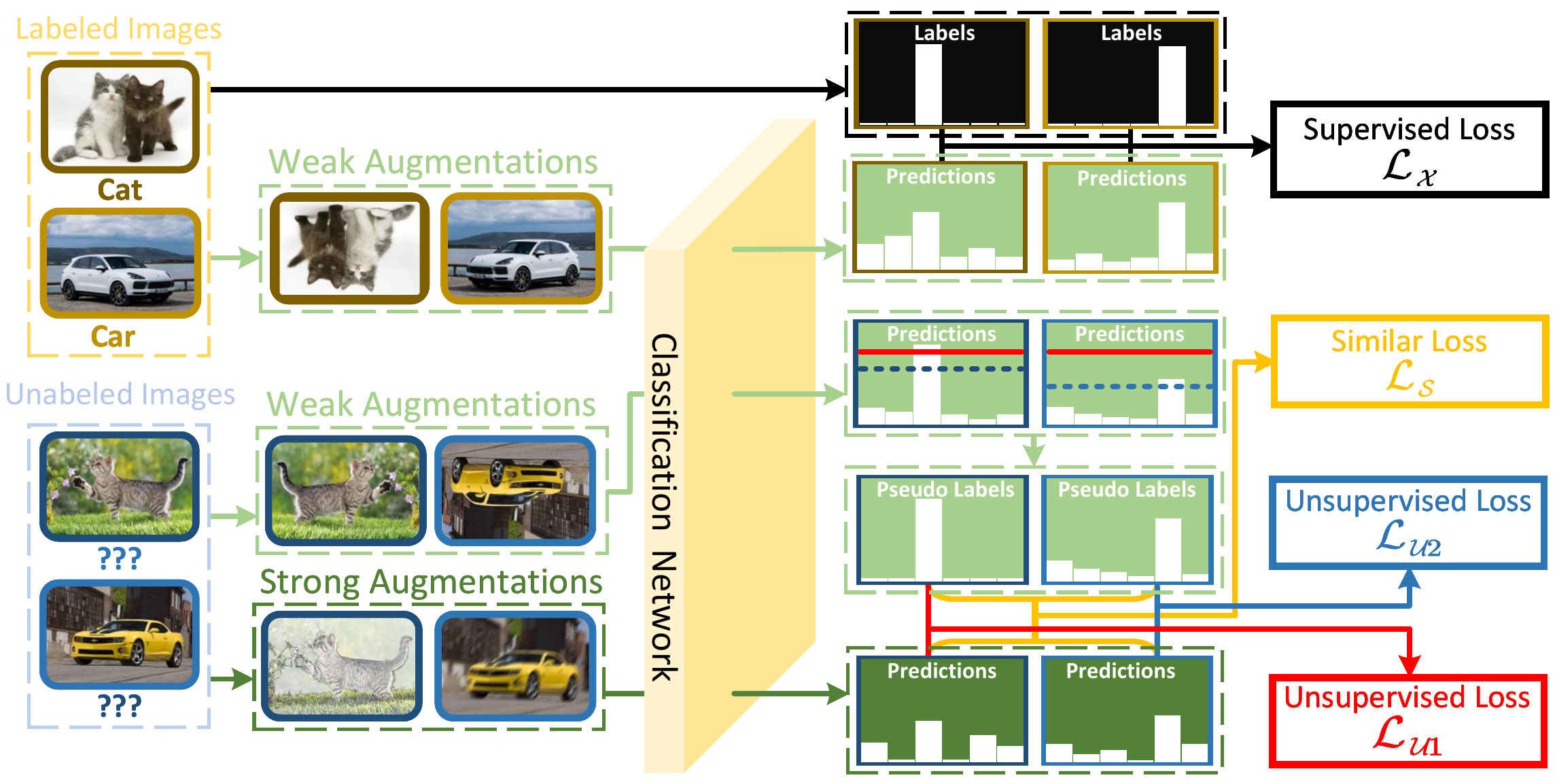}
	\caption{Diagram of the proposed ADT-SSL which is mainly divided into three parts: 1) Supervision loss $\mathcal{L} _{\mathcal{X}}$. 2) Two unsupervised losses $\mathcal{L} _{\mathcal{U} 1}$ and $\mathcal{L} _{\mathcal{U} 2}$ focus on the consistency of the model's predictions for the strongly and weakly augmented versions of the same unlabeled data, respectively. 3) Similar loss $\mathcal{L} _{\mathcal{S}}$ considers the consistency of the model's prediction of highly similar unlabeled data.}\label{Fig1}
	\label{fig1}
\end{figure}

\section{RELATED WORK}
To make this paper self-contained, we briefly describe several basic backgrounds.

\noindent
\textbf{Consistency Regularization:}
Consistency regularization was first proposed in \cite{laine2016temporal, sajjadi2016regularization} and then developed in \cite{sohn2020fixmatch, berthelot2019remixmatch}. Its core idea is that when different disturbed versions of the same data are input to the model, their predictions should remain as consistent as possible. Consistency regularization has been widely used in SSL since it can efficiently exploit the unlabeled data. This technique can be implemented by minimizing the following loss:
\begin{equation}\label{eq1}
\left\| p_m\left( y|D\left( x \right) ;\theta \right) -p_m\left( y|D\left( x \right) ;\theta \right) \right\| _{2}^{2},
\end{equation}
where $D\left( x \right)$ and $y$ denote a random disturbance to the input data $x$ and the output of the model $p_m\left( \cdot|\cdot \,;\cdot \right)$, respectively. Note that the two terms in Eq. (\ref{eq1}) may be different due to the randomness of the function $D\left( \cdot \right)$. 


\noindent
\textbf{Data Augmentation and Anchoring:}
As network becomes deeper, tremendous data are required for training a good model. In practice, the labeled data are not much enough and the quality of dataset is not good enough. Data augmentation was proposed to alleviate this problem by deforming image or adding disturbance to the image without changing the image semantics to extend the dataset. In general, augmentation way used in SSL includes weak augmentation and strong augmentation. Weak augmentation involves image flipping, cutting, translation, etc. By contrast, strong augmentation has stronger disturbances, such as random affine transformation and color jitter \cite{berthelot2019remixmatch, cubuk2019autoaugment, cubuk2020randaugment}.

By combining consistency regularization and data augmentation, augmentation anchoring was first proposed in ReMixMatch \cite{berthelot2019remixmatch}. It first weakly and strongly augments the unlabeled data, separately, and then sets the output of the model to the weakly augmented data as an ``anchor''. For the same unlabeled data, we hope that the model's prediction of the strongly augmented version is consistent with the weakly augmented one, which can be implemented by minimizing
\begin{equation}
\left\| p_m\left( y|w\left( u \right) ;\theta \right) -p_m\left( y|\mho \left( u \right) ;\theta \right) \right\| _{2}^{2},
\end{equation}
\noindent
where $w\left( u \right)$ and $\mho \left( u \right)$ denote the weak augmentation and strong augmentation of the input data $u$, respectively. Augmentation anchoring increases the stability of consistency regularization by invoking different strengths of data augmentation method.

\begin{algorithm}[tp]
    \begin{algorithmic}[1]  
	\caption{Class-Adaptive Threshold Extraction Algorithm.}  
    \label{alg1} 
    \State \textbf{Input:} $E$ and $I$ denote the numbers of epochs and iterations, respectively. $\mathcal{X}=\{\left( x_b,y_b \right)\}$ denotes a batch of labeled samples and their one-hot labels, where $b=1,...,B$.
	\For{$e$=1 to E}
		\State $\mathcal{T} _{c}^{0}$ = 0.95  \textcolor{gray}{\emph{\ \ \ \ \ \ \ \ \ \ \ \ \ \ \ \ \ \ \ \ \ \ \ \ \ \ \ \ \ \ \ \ \ \ \ \ \ \ \ \ \ \ \ // Reassign a value to $\mathcal{T} _{c}^{0}$ at the beginning of each Epoch.}}
		\For{$i=1$ to $I$}
			\For{$b$=1 to $B$}
				\State $c=\textrm{arg}\max \left( y_b \right)$ \textcolor{gray}{\emph{\ \ \ \ \ \ \ \ \ \ \ \ \ \ \ \ \ \ \ \ \ \ \ \ \ \ \ \ \ \ \ \ \ // c represents the category to which $x_b$ belongs.}}
				\State $q=p_m\left( \tilde{y}|w\left( x_b \right) ;\theta \right) $ \textcolor{gray}{\emph{\ \ \ \ \ \ \ \ \ \ \ \ \ \ \ \ // Output model predictions for weakly augmented $x_b$.}}
				\If{$\textrm{arg}\max \left( q \right) ==c\,\, \textrm{and}\,\,\max \left( q \right) <\mathcal{T} _c$}
					\State $\mathcal{T} _c=\max \left( q \right)$ \textcolor{gray}{\emph{\ \ \ \ \ \ \ \ \ \ \ \ \ \ \ // Adjust the class adaptive threshold corresponding to $x_b$.}}
				\EndIf
				\If{$\textrm{arg}\max \left( q \right) ==c\,\,\textrm{and}\,\,\max \left( q \right) <\mathcal{T} _{c}^{0}$}
					\State $\mathcal{T} _{c}^{0}=\max \left( q \right)$ \textcolor{gray}{\ \ \ \ \ \ \ \ \ \ \ \emph{// Since $\mathcal{T} _{c}^{0}$ will be recalculated, $\mathcal{T} _{c}^{0}$ and $\mathcal{T} _c$ may be different.}}
				\EndIf
			\EndFor
		\EndFor 
		\If{$\mathcal{T} _{c}^{0}>\mathcal{T} _c$} 
			\State $\mathcal{T} _c=\mathcal{T} _{c}^{0}$ \textcolor{gray}{\emph{\ \ \ \ \ \ \ \ \ \ \ \ \ \ \ \ \ \ \ \ // At the end of each epoch, we consider whether to assign $\mathcal{T} _{c}^{0}$ to $\mathcal{T} _c$.}}
		\EndIf
	\EndFor 
  \end{algorithmic}  
\end{algorithm}

\noindent
\textbf{Pseudo-Labeling and Propagation:}
Pseudo-labeling is to reassign a label to the unlabeled data for the dataset extension, whose loss term can be written as
\begin{equation}
\mathbb{1}\left( \max \left( q \right) >\tau \right) \cdot \textrm{H}\left( \textrm{one}\left(q \right) ,p_m\left( y|\mho \left( u \right) ;\theta \right) \right), 
\end{equation}
where $q=p_m\left( y|w\left(u\right);\theta \right)$ for abbreviation, $\tau$ denotes the fixed threshold, $\textrm{one}(\cdot)$ function converts the input distribution into a one-hot label, and $\textrm{H}(\cdot,\cdot)$ denotes the CE loss. It uses the model trained on the labeled data to predict the unlabeled data. If the prediction is above the threshold $\tau$, we assign the predicted label to the unlabeled data. This predicted label is called the pseudo label of an unlabeled data. Hence, we can input the pseudo-labeled data to train a model. Based on the assumption of low-density separation \cite{chapelle2009semi}, the data density at the decision boundary should be as low as possible. To this end, the authors in \cite{sohn2020fixmatch, hu2021simple} set a very high fixed confidence threshold, e.g., $\tau=0.95$, which makes the model focus on the unlabeled data that are far from the decision boundary \cite{lee2013pseudo}.

	Label propagation was first proposed in graph-based SSL \cite{chapelle2009semi}. The basic idea is to predict the label information of unlabeled nodes from the labeled nodes. In \cite{douze2018low}, the authors proposed to select the closest node to construct the graph. Inspired by \cite{douze2018low}, the methods \cite{hu2021simple, iscen2019label} use this technology to set pseudo labels for unlabeled data. When the similarity between the unlabeled data is high enough, the label may be spread from an unlabeled data to another one. As done in \cite{hu2021simple, iscen2019label}, this paper uses the same similar threshold to control the label propagation.

\section{Proposed Method}
To boost the performance of SSL, we first propose an Adaptive Dual-Threshold (ADT) method, which can extract more useful information from the unlabeled data. For highly similar unlabeled data, we then design a novel similar loss to learn the consistency of the model. Finally, our overall loss consists of four different loss functions, which serve four different types of data, respectively. Figure \ref{fig1} gives an overview of the proposed ADT-SSL.
 
\subsection{Mathematical Model}
Denote a batch of labeled data as $\mathcal{X} =\{\left( x_b,y_b \right)\}$ and a batch of unlabeled data as $\mathcal{U} =\{u_b\}$, $b=1,...,B$. $C$ denotes the total number of classes of a dataset. Augmentation anchoring \cite{berthelot2019remixmatch} is used in our algorithm. The model's prediction on the weakly augmented unlabeled data is used as the anchor of the strongly augmented unlabeled data. Like FixMatch \cite{sohn2020fixmatch}, MixMatch \cite{berthelot2019mixmatch}, and ReMixMatch \cite{berthelot2019remixmatch}, the weak augmentation on CIFAR-10 and CIFAR-100 is to randomly flip an image with a probability of 50$\%$ and then randomly translate the image horizontally or vertically by 12.5$\%$. We also apply RandAugment \cite{cubuk2020randaugment} to perform strong augmentation on unlabeled data. RandAugment contains a variety of transformations, and its diversity is determined by predefined parameter.

\begin{figure}[tp]
  \subfloat[]{\label{a} \includegraphics[width=0.53\textwidth]{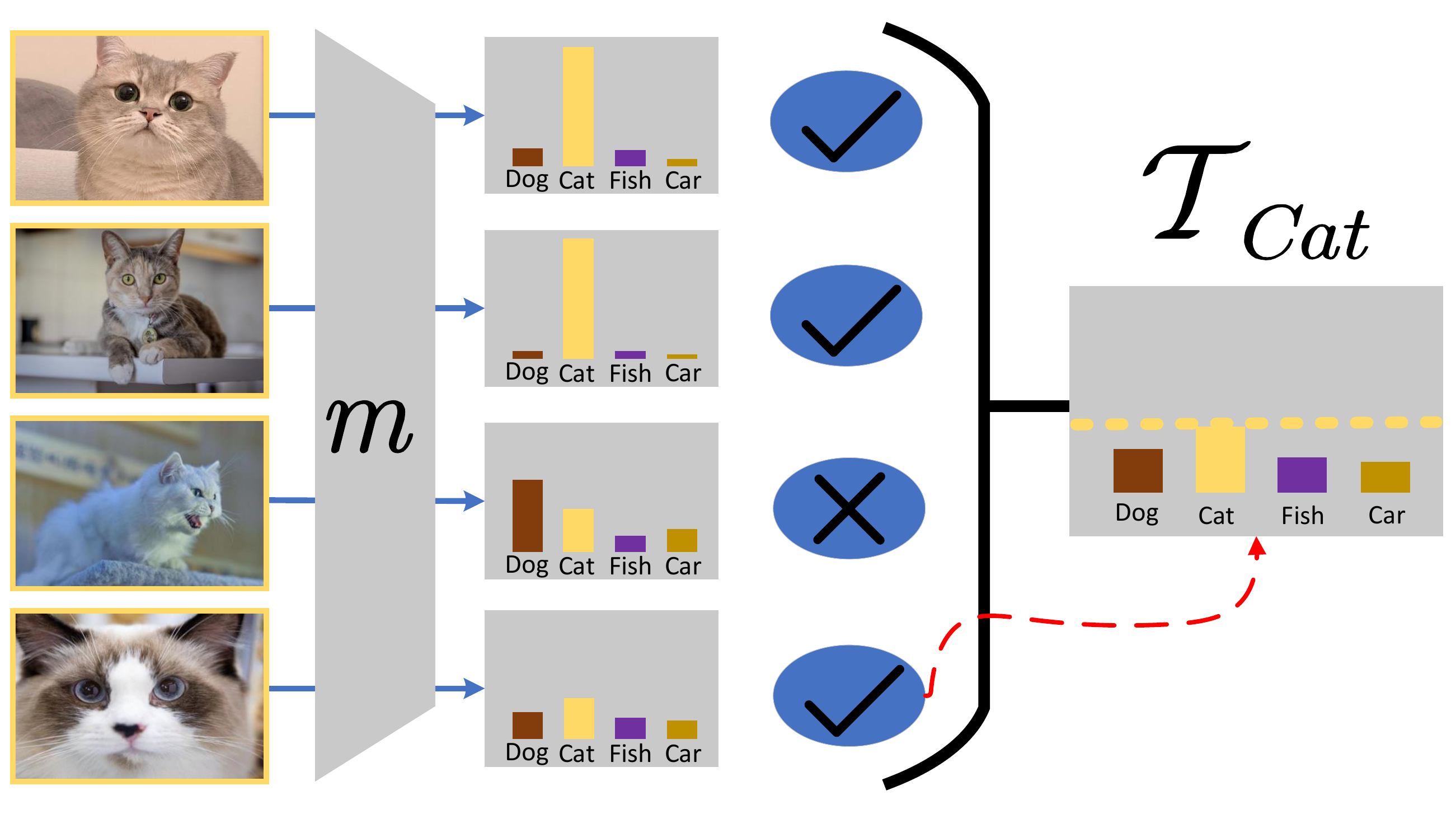}}
 \hfill 	
  \subfloat[]{\label{b} \includegraphics[width=0.47\textwidth]{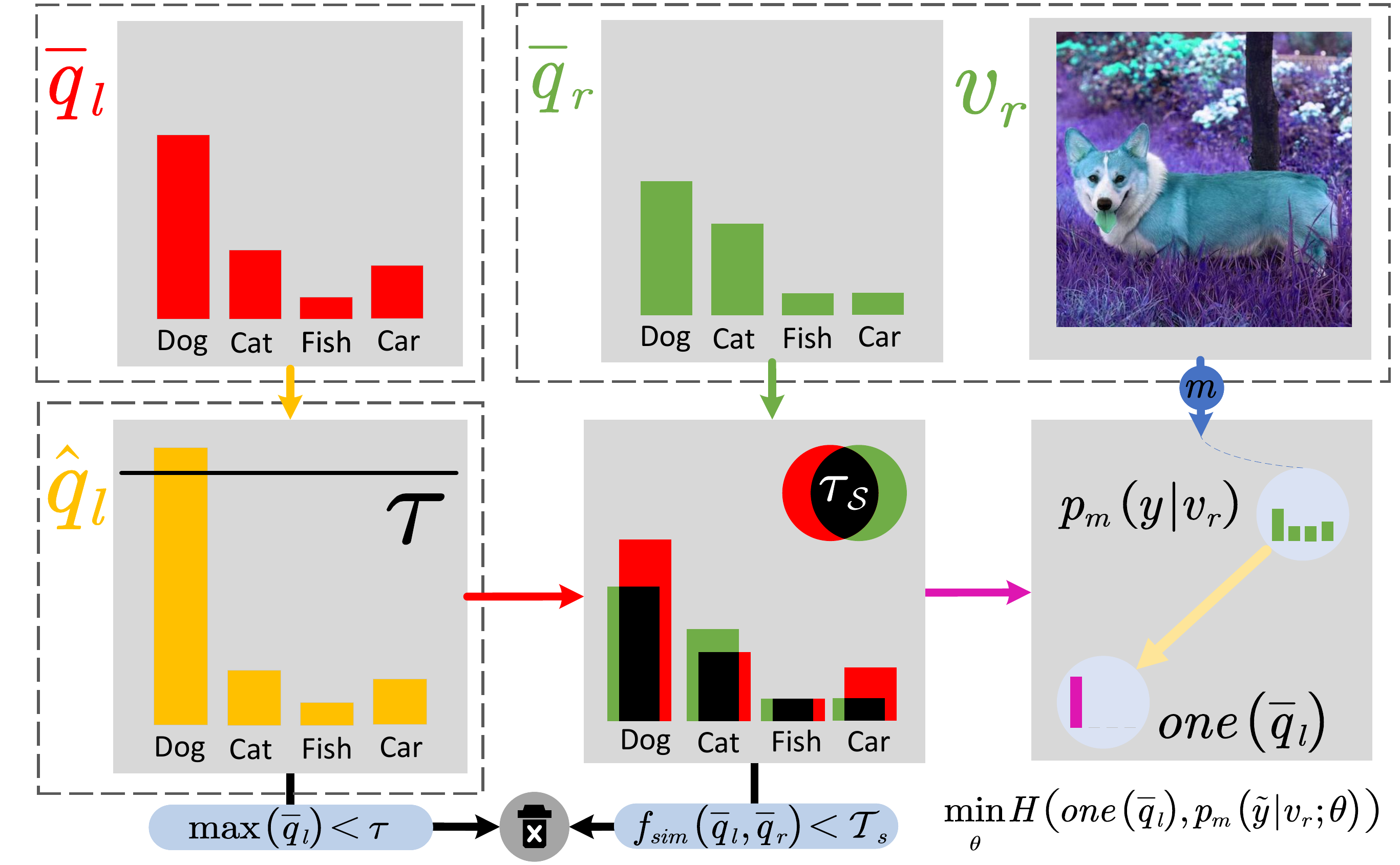}}
  \vspace{-0.5em}
\caption{\protect\subref{a} Illustration of extracting class-adaptive threshold. For a certain class (e.g., ``cat''), we take the smallest prediction value of this class as the adaptive threshold of the class among all the samples of this class which can be correctly classified by the model. As shown in the rightmost block, the yellow dashed-line represents the adaptive threshold of the class ``Cat''.
\protect\subref{b} Expatiate of similar loss. Given a distribution $\bar{q}_l$ which is the model's prediction of a weakly augmented unlabeled data, we first perform a sharpen operation on $\bar{q}_l$ to get $\hat{q}_l$. Then we predict its confidence. When the maximum value of $\hat{q}_l$ is higher than the confidence threshold $\tau$, we convert it to the one-hot label and use it as an ``anchor''. Then we will evaluate the similarity of $\bar{q}_l$ to another unlabeled data tuple $(\bar{q}_r, v_r)$. When the similarity is greater than the similarity threshold $\mathcal{T} _s$, we use $\textrm{one}\left( \hat{q}_l \right)$ as the pseudo label of $p_m\left( y|v_r \right)$ to calculate the cross entropy loss in order to reduce the distance between $v_r$ and its corresponding clustering center, where $v_r$ is the strongly augmented vision of other unlabeled data. It is worth noting that Similar Loss on $\bar{q}_l$, $\bar{q}_r$, $v_r$ will not work when the two thresholds cannot be exceeded simultaneously.}
\label{fig2}
\end{figure}
\subsection{Class-Adaptive Threshold}
The current SSL methods set the same fixed threshold for all the classes to select the unlabeled data for the model training \cite{sohn2020fixmatch, hu2021simple}. Since the fixed threshold is very large (e.g., 0.95), some hard samples do not have the opportunity to join the model training. This makes the model unable to learn well for those categories which are difficult to be recognized. Our ADT can adaptively adjust the learning strategy based on the model's learning situation for different classes at different training stages. This process can be implemented by extracting the class-adaptive thresholds from the labeled data.

\noindent
\textbf{Adaptive Threshold Extraction:} Assume that the learning state of the model can be reflected by its prediction of labeled data. For labeled data, we perform the following operation:
\begin{equation}
\begin{split}
\textrm{if}\,\,\textrm{arg}&\max \left( p_m\left( \tilde{y}|w\left( x \right) ;\theta \right) \right) ==c\,\,\textrm{and}\,\max \left( p_m\left( \tilde{y}|w\left( x \right) ;\theta \right) \right) < \mathcal{T} _c:
\\
&\mathcal{T} _c=\max \left( p_m\left( \tilde{y}|w\left( x \right) ;\theta \right) \right),
\end{split}\label{Adapt}
\end{equation}
where $c$ denotes the class label which $x$ belongs to and $\mathcal{T} _c$ is the adaptive threshold of the class $c$ where its initial value is set to 0.95. By Eq. (\ref{Adapt}), our method first determines whether the model's prediction of $w\left( x \right)$ is correct, and then performs the comparison operation to check whether the maximum value predicted by the model is less than $\mathcal{T} _c$. Only the two conditions of Eq. (\ref{Adapt}) hold, $\max \left( \tilde{y} \right)$ at this time will be assigned to $\mathcal{T} _c$. In this way, we can set the class-adaptive thresholds for all the classes. Figure \ref{a} illustrates the process of the class-adaptive threshold extraction.

\noindent
\textbf{Algorithm Implementation:} Since the SSL methods simultaneously exploit the labeled and unlabeled data to train the model,  the extraction algorithm will be executed in the whole training process. Note that $\mathcal{T} _c$ does not always have an increasing or decreasing trend during the training process. Hence, for each epoch we will recalculate a new $\mathcal{T} _{c}^{0}$ for each class and assign $\mathcal{T} _{c}^{0}$ to $\mathcal{T} _c$ if $\mathcal{T} _{c}^{0}$ is larger than $\mathcal{T} _c$. Interestingly, our algorithm does not need to introduce additional validation set to evaluate the learning state of the model at each epoch, and only needs to record the model output of the labeled data. As a result, our method achieves an additional gain from learning on labeled data without increasing the overall computational budget. The complete process of extracting the class-adaptive threshold is shown in Algorithm \ref{alg1}.

In order to extract more effective information from the unlabeled data and balance the learning state of each class, we combine the high fixed threshold $\tau$ and class-adaptive threshold $\mathcal{T} _c$ to form an Adaptive Dual-Threshold. When confirming a pseudo label, the dual-threshold structure can screen the unlabeled data more carefully. And we can design different training methods for these two types of unlabeled data. Inspired by entropy minimization, we use sharpen operation \cite{berthelot2019mixmatch} to reduce the entropy of pseudo label distribution in the process of generating pseudo labels. This technology has also been widely used in other popular SSL algorithms \cite{berthelot2019remixmatch, hu2021simple}, which can be expressed as:
\begin{equation}
\textrm{Sharpen}\left( p,T \right) =p_{c}^{1/T}(\sum\nolimits_{c=0}^{C-1}{p_{c}^{1/T}})^{-1},
\end{equation}
where $p$ represents the distribution of the input and $T$ is a predefined hyper parameter ranging from 0 to 1. This function is to adjust the sharpness of the input. 
When $T\rightarrow 0$, the output of the function will be close to the one-hot distribution. Next we use it to make the model output a low entropy result.

Let $q_b=p_m\left( \tilde{y}|w\left( u_b \right) ;\theta \right)$ represent the model output of $w\left( u_b \right)$ and denote $\bar{q}_b$ as an Exponential Moving Average (EMA) on $q_b$. Compared with the direct output of the model to the unlabeled data, the distribution obtained by EMA is smoother and the prediction will not fluctuate greatly due to a certain abnormal value. For the unlabeled data, we may rely on the fixed threshold $\tau $ to determine whether the following loss needs to be calculated:
\begin{equation}
\mathbb{1}\left( \max \left( \hat{q}_b > \tau \right) \right) \cdot \textrm{H}\left(\textrm{one}\left( \hat{q}_b \right) ,p_m\left( \tilde{y}|\mho \left( u_b \right) ;\theta \right) \right),  
\end{equation}
where $\hat{q}_b=\textrm{Sharpen}\left(\bar{q}_b,T \right)$. In classification tasks, if the confidence of the input unlabeled data is high enough, the model can benefit by using the one-hot label to calculate the CE loss of the input unlabeled data. This is similar to the learning from labeled data. If the maximum value of $\hat{q}_b$ exceeds $\tau$, it means that the unlabeled data has a high enough confidence level. Then, we will convert $\hat{q}_b$ to one-hot label to calculate the CE loss. In existing SSL methods, $\tau$ is often a high fixed threshold for all the classes. In this paper, we propose to use the class-adaptive threshold $\mathcal{T} _c$ to rescreen the unlabeled data whose predictions are less than $\tau$, and then adopt $L_2$-norm loss learning strategy:
\begin{equation}
\mathbb{1}\left( \max \left( \hat{q}_b \right) <\tau \right) \cdot \mathbb{1}\left( \max \left( \bar{q}_b \right) >\mathcal{T} _{\textrm{arg}\max \left( \bar{q}_b \right)} \right) \cdot \left\| \hat{q}_b-p_m\left( \tilde{y}|\mho \left( u_b \right) ;\theta \right) \right\| _{2}^{2},
\end{equation}
where $\mathcal{T}_{\textrm{arg}\max \left( \bar{q}_b \right)}$ represents the class-adaptive threshold corresponding to the maximum value of $\bar{q}_b$. In order to make the selection of the unlabeled data more accurate, we replace $\hat{q}_b$ with $\bar{q}_b$ to determine whether $u_b$ is added to the training process. Once this unlabeled data satisfies $\max \left( \hat{q}_b \right) <\tau$ and $\max \left( \bar{q}_b \right) >\mathcal{T}_{\textrm{arg}\max \left( \bar{q}_b \right)}$, we use $L_2$-norm to compute the loss between $\hat{q}_b$ and $\tilde{y}$. Taking $\hat{q}_b$ as the pseudo label, we can achieve not only the minimum entropy,  but also stronger generalization ability to the model because the soft label $\hat{q}_b$ carries more information. Most importantly, the soft label is not as sensitive to noisy data as one-hot label.

\subsection{Similar Loss}
To make full use of the unlabeled data, we propose a novel similar loss to capture the consistency information between the different unlabeled data. In this process, we allow pseudo labels to propagate between the highly similar unlabelled data. We take a pseudo label of weakly augmented unlabeled data with high confidence as the ``anchor'' and then align the strongly augmented unlabeled data that is sufficiently similar to the weakly augmented unlabeled data with the ``anchor''. We show the detailed process in Figure \ref{b} and define the similar loss as follows:
\begin{equation}
\begin{aligned}
\mathcal{L} _{\mathcal{S}}=\frac{1}{\left( \begin{array}{c}
	K'B\\
	2\\
\end{array} \right)}\sum_{\begin{array}{c}
	i,j\epsilon [|\mathcal{U}'|], i\ne j\\
	\left( v_l,\bar{q}_l,\hat{q}_l \right) =\mathcal{U} '_i\\
	\left( v_r,\bar{q}_r,\hat{q}_r \right) =\mathcal{U} '_j\\
\end{array}}{\mathbb{1}\left( \max \left( \hat{q}_l \right) >\tau \right)}&\cdot \mathbb{1}\left(\textrm{Sim}\left( \bar{q}_l,\bar{q}_r \right) >\mathcal{T} _s \right) 
\\
&\cdot \textrm{H}\left(\textrm{one}\left( \bar{q}_l \right) ,p_m\left( \tilde{y}|v_r;\theta \right) \right),  
\end{aligned}
\end{equation}
where $\textrm{Sim}\left(\cdot,\cdot\right)=\sqrt{\cdot}^{\top}\sqrt{\cdot}$ is the similarity function measured by the Bhattacharyya distance \cite{bhattacharyya1946measure} and $\mathcal{T} _s$ is the similarity threshold between the different unlabeled data. The Bhattacharyya distance measures the degree of overlap between two distributions, allowing us to choose the similarity threshold ($\mathcal{T} _s$) more intuitively. Noteworthy, we use $\hat{q}_l$ when determining whether the unlabeled data exceeds the confidence threshold ($\tau$). For the input distributions of the $\textrm{Sim}\left(\cdot,\cdot\right)$ function, we use $\bar{q}_l$ and $\bar{q}_r$ (without sharpen operation) to remain closer to the original output of the model. The purpose is to make the similarity calculation between two unlabeled data more accurate.

In this paper, we set $\tau=0.95$ and $\mathcal{T} _s=0.9$. At this time, the distribution of unlabeled data that exceeds the high confidence threshold ($\tau$) will be close to the one-hot distribution, and the other unlabeled data whose similarity is larger than the similarity threshold ($\mathcal{T} _s$) also have a high confidence. This means that we can propagate a pseudo label to $v_r$ with high confidence. Therefore, when using $v_r$ to train the modal, we take one-hot label as the target of $p_m\left( \tilde{y}|v_r;\theta \right) $ to calculate the CE loss.
\begin{algorithm}[tp]  
    \begin{algorithmic}[1]
    \caption{Proposed ADT-SSL method.}
    \label{alg2}  
    \State \textbf{Input:} Batch of labeled samples and their one-hot labels $\mathcal{X}=\{\left( x_b,y_b \right)\}$,  batch of unlabeled samples $\mathcal{U} =\{u_b\}$, $b=1,...,B$, sharpening temperature $T$, number of weak augmentations $K$, number of strong augmentations $K_{strong}$, fixedthreshold $\tau$, similarity threshold $\mathcal{T}_s$, class-adaptive threshold $\mathcal{T} _c$.  
    \For{$b$=1 to $B$}
        \State $\tilde{x}_b=w\left( x_b \right)$ \textcolor{gray}{\emph{\ \ \ \ \ \ \ \ \ \ \ \ \ \ \ \ \ \ \ \ \ \ \ \ \ \ \ \ \ \ \ \ \ \ \ \ \ \ \ \ \ \ \ \ \ \ // Implementing weak augmentation for labeled data.}}
        \For{$k$=1 to $K$}
				\State $\tilde{u}_{b,k}=w\left( u_b \right)$ \textcolor{gray}{\ \ \ \ \ \ \ \ \ \ \ \ \ \ \ \ \ \ \emph{// Implement $K$ kinds of weak augmentation for unlabeled data.}}
			\EndFor
			\For{$k$=1 to $K_{strong}$}
				\State $\hat{u}_{b,k}=\mho\left( u_b \right)$ \textcolor{gray}{\emph{\ \ \ \ \ \ // Implement  $K_{strong}$ kinds of strong augmentation for unlabeled data.}}
			\EndFor
			\State $\bar{q}_b=\frac{1}{K}\sum\nolimits_{k=1}^K{p_m\left( \tilde{y}\,\,| \tilde{u}_{b,k};\theta \right)}$ \textcolor{gray}{\emph{\ \ // Calculate the average prediction  for weakly augmented $u_b$.}}
			\State $\hat{q}_b=\textrm{sharpen}\left( \bar{q}_b,T \right)$ \textcolor{gray}{\ \ \ \ \ \ \ \ \ \ \ \ \ \emph{// \textrm{Sharpen} operation is applied to the average prediction results.}}  			
  	 \EndFor
	   \State $\tilde{\mathcal{X}}=\{\left( \tilde{x}_b,y_b \right)\}; b=1,...,B$ \textcolor{gray}{\ \ \ \ \ \ \ \ \ \ \ \ \ \ \ \ \ \ \ \ \ \ \ \ \ \ \ \ \ \ \ \ \ \emph{//Weakly augmented label data and their labels.}}
		\State $\hat{\mathcal{U}}=\{\left( \hat{u}_{b,k},\bar{q}_b,\hat{q}_b \right)\};b=1,...,B$ \textcolor{gray}{\emph{\ \ \ \ \ \ //Strongly augmented unlabel data and their pseudo labels.}}
		\State Calculate $\mathcal{L} _{\mathcal{X}}$, $\mathcal{L} _{\mathcal{U} 1}$, $\mathcal{L} _{\mathcal{U} 2}$, and $\mathcal{L} _{\mathcal{S}}$ by Eqs. (10), (11), (12), and (8), respectively. 

		\State\textbf{Return:} $\mathcal{L} _{\mathcal{X}}+\lambda _{\mathcal{U} 1}\mathcal{L} _{\mathcal{U} 1}+\lambda _{\mathcal{U} 2}\mathcal{L} _{\mathcal{U} 2}+\lambda _{\mathcal{S}}\mathcal{L} _{\mathcal{S}}$  \textcolor{gray}{\emph{\ \ \ \ \ \ \ \ \ \ \ \ \ \ \ \ \ \ \ \ \ \ \ \ \ \ \ \ \ \ \ \ \ \ \ \ \ \ \ \ \ \ \ \ \ \ // Calculate overall loss.}} 
  \end{algorithmic}
\end{algorithm}

\subsection{Overall Loss}
Combining the proposed similar loss, our overall loss is composed of four components, which are supervised loss $\mathcal{L} _{\mathcal{X}}$, two unsupervised losses $\mathcal{L} _{\mathcal{U} \,1}$, $\mathcal{L} _{\mathcal{U} \,2}$, and similar loss $\mathcal{L} _{\mathcal{S}}$:
\begin{equation}
\mathcal{L} =\mathcal{L}_{\mathcal{X}}+\lambda _{\mathcal{U} 1}\mathcal{L}_{\mathcal{U} 1}+\lambda_{\mathcal{U} 2}\mathcal{L} _{\mathcal{U} 2}+\lambda_{\mathcal{S}}\mathcal{L}_{\mathcal{S}},
\end{equation}
where
\vspace{-1em}
\begin{equation}
\mathcal{L} _{\mathcal{X}}=\frac{1}{\left| \mathcal{X} ' \right|}\sum_{x,y\epsilon \hat{\mathcal{X}}}{\textrm{H}\left( y;p_m\left( \tilde{y}|x;\theta \right) \right)},
\end{equation}
\begin{equation}
\mathcal{L} _{\mathcal{U} 1}=\frac{1}{| \hat{\mathcal{U}}|}\sum_{u,q\epsilon \hat{\mathcal{U}}}{\mathbb{1}\left( \max \left( \hat{q}> \tau \right) \right) \cdot \textrm{H}\left(\textrm{one}\left( \hat{q} \right) ,p_m\left( \tilde{y}|u;\theta \right) \right)},
\end{equation}
\noindent
and
\begin{equation}
\mathcal{L} _{\mathcal{U} 2}=\frac{1}{L| \hat{\mathcal{U}}|}\sum_{u,q\epsilon \hat{\mathcal{U}}}{\mathbb{1}\left( \max \left( \hat{q}<\tau \right) \right) \cdot \mathbb{1}\left( \bar{q}>\mathcal{T} _c \right) \cdot \left\| \hat{q}-p_m\left( \tilde{y}|u;\theta \right) \right\| _{2}^{2}},
\end{equation}
where $\mathcal{L} _{\mathcal{X}}$ is the CE loss calculated for the weakly augmented labeled data and $\vert \cdot \vert$ denotes the set cardinality. Two super parameters $\mathcal{L} _{\mathcal{U} 1}$ and $\mathcal{L} _{\mathcal{U} 2}$ focus on the consistency of two different augmentations of the same unlabeled data. For a batch of labeled and unlabeled data, we use ADT to determine the pseudo label for the unlabeled data, and apply the augmentation anchoring technology. Finally, using our similar loss drives the prediction of highly similar data tend to be consistent. The complete process of our ADT-SSL is shown in Algorithm \ref{alg2}.

\section{EXPERIMENTS}
\subsection{Experimental Setup}
\noindent
\textbf{Benchmark Datasets:} \textbf{CIFAR-10} \cite{krizhevsky2009learning}: It contains 60,000 images with 32 $\times$ 32 size and 10 classes. Each class has 6,000 samples. The test set contains 10,000 images and the training set contains 50,000 images. \textbf{CIFAR-100} \cite{krizhevsky2009learning}: Different from CIFAR-10, it contains 100 classes, and each class has only 600 samples. Except for this, the other attributes of CIFAR-100 are the same as that of CIFAR-10. \textbf{SVHN} \cite{netzer2011reading}: In SVHN, the training set contains 73,257 images and test set contains 26,032 images, all of which are 32 $\times$ 32 size. SVHN has 10 classes as well. By convention, we set the size of the validation set to 5,000 in all the experiments.

\noindent
\textbf{Baseline Methods:}
Nine baseline methods including SimPLE \cite{hu2021simple}, FixMatch \cite{sohn2020fixmatch}, $\Pi$-Model \cite{rasmus2015semi}, Pseudo-Labeling \cite{lee2013pseudo}, Mean Teacher \cite{tarvainen2017mean}, MixMatch \cite{berthelot2019mixmatch}, UDA \cite{xie2020unsupervised}, ReMixMatch \cite{berthelot2019remixmatch}, and VAT\cite{miyato2018virtual}, are added to the performance comparison. For fair comparison, all the compared methods adopt the same augmentation methods \cite{cubuk2020randaugment}, the same number of augmentation forms, and the same backbone.

\noindent
\textbf{Implementation Details:}
In the experiment, Wide ResNet 28-2 \cite{zagoruyko2016wide} with 1.5M parameters and Wide ResNet 28-8 \cite{zagoruyko2016wide} with 23M parameters are adopted as the backbones, which are abbreviated by WRN28-2 and WRN28-8 respectively for convenience. For all the experiments, we use SGD as the optimizer where the weight decay is set to 0.0005 for CIFAR-10 and SHVN, 0.001 for CIFAR-100, and the learning rate is set 0.03. Cosine learning rate decay \cite{loshchilov2016sgdr} with a decay rate of $\frac{7\pi}{16}$ is used. EMA with a decay rate of 0.999 is used for the output of the model. To verify the effectiveness of the proposed ADT-SSL, MixMatch Enhanced \cite{hu2021simple} and fully-supervised learning methods are also included for comparison. Note that MixMatch Enhanced \cite{hu2021simple} combines MixMatch and augmentation anchoring, and fully-supervised means that all labels are used for training.


\begin{figure}[ht]
  \subfloat[]{\label{c} \includegraphics[width=0.32\textwidth]{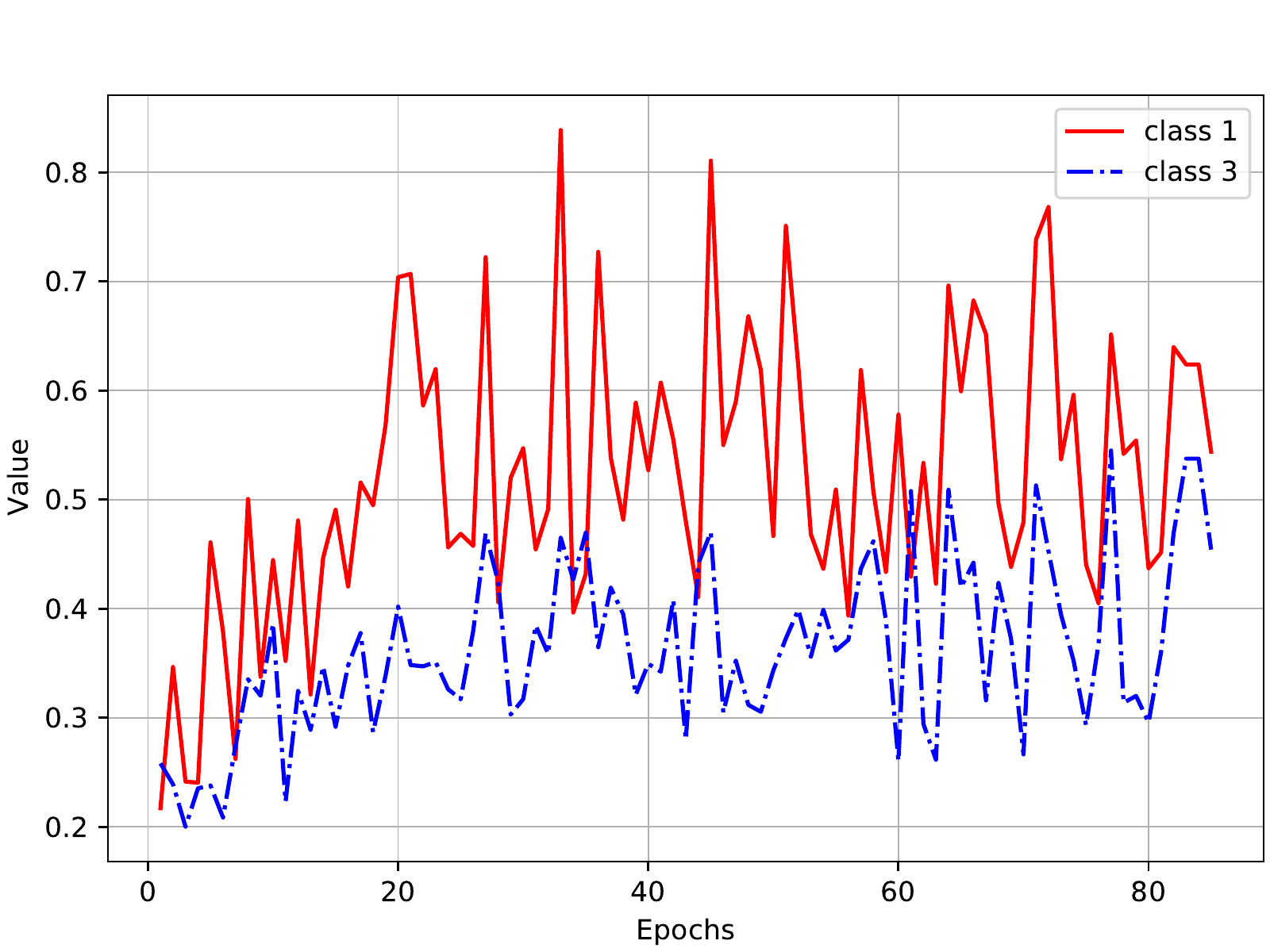}}
 \hfill 	
  \subfloat[]{\label{d} \includegraphics[width=0.32\textwidth]{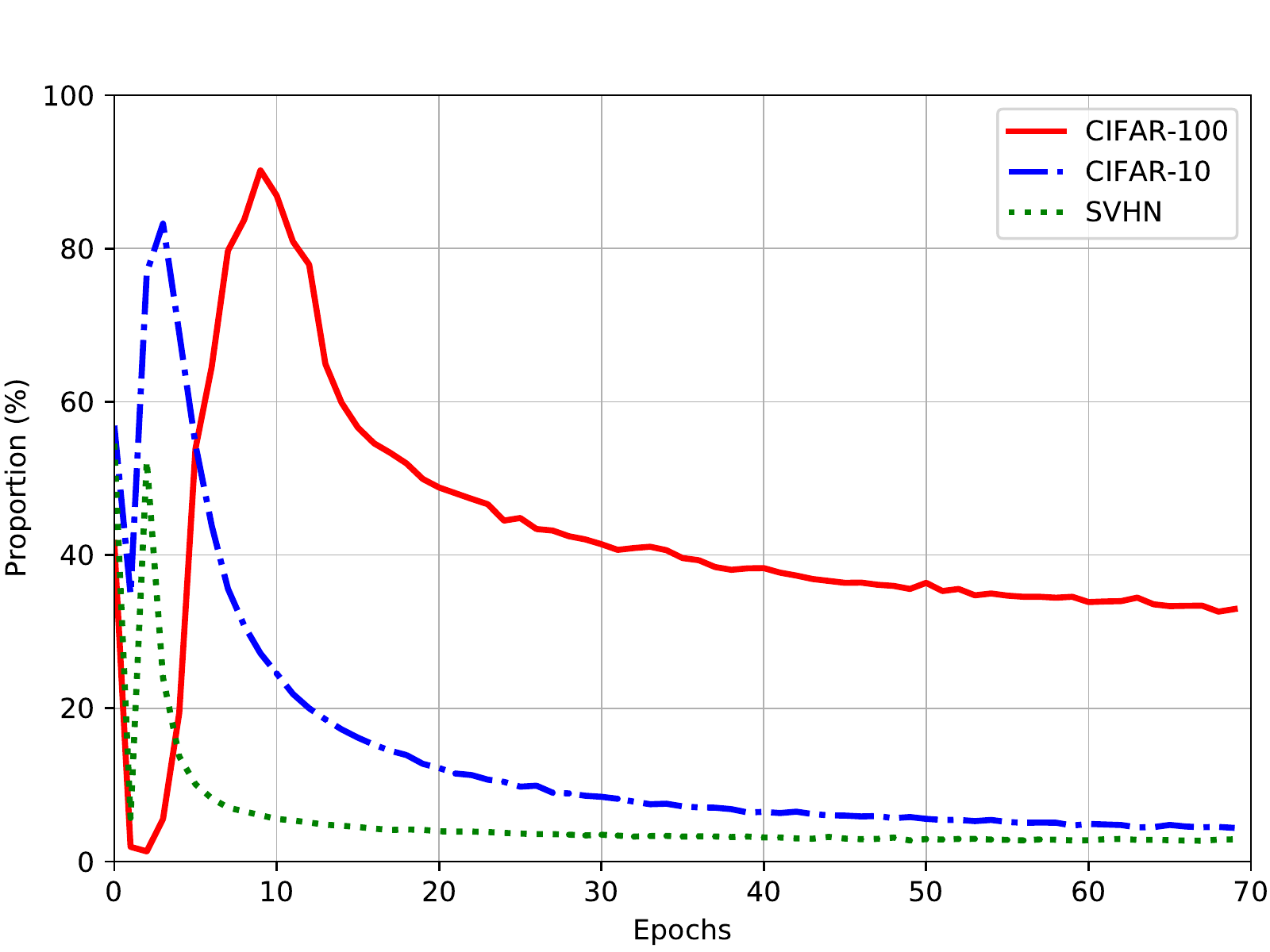}}
 \hfill	
  \subfloat[]{\label{e} \includegraphics[width=0.32\textwidth]{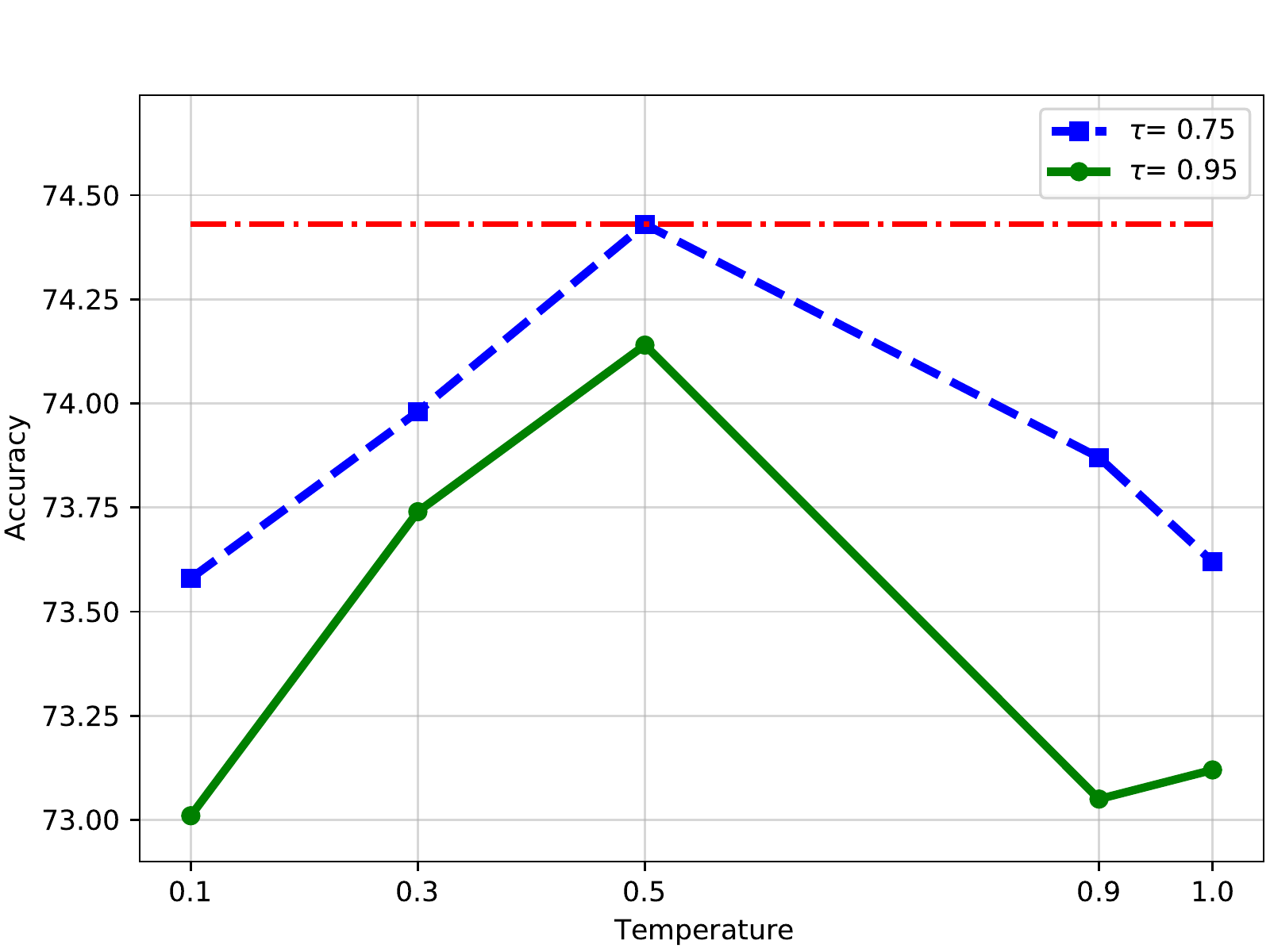}}
  \vspace{-0.5em}
\caption{\protect\subref{c} Illustration of class-adaptive threshold of CIFAR-10. 
\protect\subref{d} Illustration of the proportion of the newly mined unlabeled data.
\protect\subref{e} Illustration of classification accuracy change with temperature in sharpen operation for CIFAR-100.}
\label{fig3}
\end{figure}
\subsection{Experimental Results}
\noindent
\textbf{Adaptive Threshold and Parameter Selection:} To better illustrate our motivation of the proposed class-adaptive threshold, we conduct a threshold change experiment on CIFAR-10 with 4,000 labeled data. The result is shown in Figure \ref{c}. The red solid and blue dotted curves represent the smallest thresholds of the classes 1 and 3, respectively. We can see that the adaptive thresholds are all less than the fixed threshold (i.e. $\tau=0.95$) during first 85 epochs, and the class 3 is more difficult to learn than the class 1. Obviously, these unlabeled data whose predictions are less than 0.95 but more than the adaptive threshold can all be correctly classified. In fact, they are of hard samples and have strong ability to benefit the model training. If only one fixed threshold is used, this part of the unlabeled data will be discarded. The proposed ADT-SSL adaptively adjusts $\mathcal{T} _c$ according to the specific category and the learning state of the model. For some categories which are difficult to learn, the model will set a lower threshold $\mathcal{T}_c$ to allow more unlabeled data to participate in the training process. In this way, our method can not only balance the learning process of the model for each class, but also makes the unlabeled data better close to its corresponding clustering center. 

Besides, we show the ratio of the unlabeled data reused by ADT-SSL to the total number of unlabeled data in Figure \ref{d}. These unlabeled data can not be exploited by a single fixed threshold but can be added to the training process due to the use of ADT. We can achieve that during the first 70 epochs, both CIFAR-10 and SVHN occupy over 3\%, while CIFAR-100 occupies over 30\%. For the complex dataset (CIFAR-100), our ADT can reuse much more unlabeled data than two relatively simple datasets (CIFAR-10 and SVHN). Therefore, it is expected that the proposed ADT-SSL will show greater improvement on CIFAR-100 than on both CIFAR-10 and SVHN.

To select the optimal value of temperature ($T$) when performing sharpen operation in ADT-SSL, we test the classification accuracy on CIFAR-100 with different fixed thresholds. In this test, Wide ResNet 28-2 is used as the backbone and the number of labeled data is 10,000. Two fixed thresholds ($\tau=0.95$ and $\tau=0.75$) are tested. The accuracy change with the parameter $T$ is shown in Figure \ref{e}. We can see that when the temperature is 0.5, the model achieves the highest accuracy (i.e., the lowest entropy), which indicates ADT-SSL introduces as few noise samples as possible.
\begin{table*}[tp]
\renewcommand\arraystretch{1}
\centering
\setlength{\tabcolsep}{2.5mm}
\caption{Classification accuracy on CIFAR-10, CIFAR-100, and SHVN. Here, Wide ResNet 28-2 backbone is for CIFAR-10 and SHVN, and Wide ResNet 28-8 backbone is for CIFAR-100.}
\vspace{-0.5em}
\label{tab1}
\begin{tabular}{l|ll|l|ll}
\hline
&\multicolumn{2}{|c|}{CIFAR-10}&CIFAR-100&\multicolumn{2}{|c}{SVHN}\\
\hline
Methods &1000 labels&4000 labels&2500 labels&1000 labels&4000 labels\\
\hline
\hline
$\Pi$-Model \cite{rasmus2015semi} &80.25\% &85.99\%&42.75\%&92.46\%&92.63\%\\
Pseudo-Labeling \cite{lee2013pseudo}&81.27\% &83.91\%&42.62\%&90.06\%&90.32\%\\
Mean Teacher  \cite{tarvainen2017mean}&82.68\%&90.81\%&46.09\%&96.58\%&96.61\%\\
VAT \cite{miyato2018virtual} &81.36\%&88.95\%&51.98\%&94.02\%&95.80\%\\
MixMatch \cite{berthelot2019mixmatch} &92.25\%&93.58\%&60.06\%&96.50\%&97.11\%\\
UDA \cite{xie2020unsupervised}&94.13\% &95.12\% &66.87\%&97.45\% &97.53\% \\
ReMixMatch \cite{berthelot2019remixmatch}&94.27\%&95.28\%&72.57\%&97.35\%&97.58\%\\
FixMatch \cite{sohn2020fixmatch}&94.67\%&95.69\%  &71.71\%&\textbf{97.64}\%&\textbf{97.83\%}\\
SimPLE \cite{hu2021simple}&94.81\%&94.95\%&71.83\%&97.54\%&97.31\%\\
\textbf{Proposed ADT-SSL}&\textbf{94.82}\%&\textbf{95.74\%}&\textbf{72.97\%}&97.13\%&97.45\%\\
\hline
Fully-Supervised&\multicolumn{2}{|c|}{95.75\%}&80.93\%&\multicolumn{2}{|c}{97.30\%}\\
\hline
\end{tabular}
\end{table*}

\noindent
\textbf{Results on CIFAR-10:} For CIFAR-10, we empirically set $\lambda _{\mathcal{U} 1}=3$, $\lambda _{\mathcal{U} 2}=225$, and $\lambda _{\mathcal{S}}=16$. The results on 1,000 and 4,000 labeled data are shown in Table \ref{tab1}. We can see from the second and third columns of Table \ref{tab1} that our ADT-SSL obtains the best performance. Note that CIFAR-10 has 60,000 labeled data. To our surprise, using only 4,000 labeled data (about 6.7\% of the total labeled data), the proposed ADT-SSL achieves almost the same performance as the fully-supervised baseline. The difference in classification accuracy is only 0.01\%. It can also be observed that all the compared methods perform better with increasing the number of labeled data from 1,000 to 4,000.

\noindent
\textbf{Results on CIFAR-100:} For CIFAR-100, we empirically set $\lambda _{\mathcal{U} 1}$ = 1, $\lambda _{\mathcal{U} 2}$ = 150, and $\lambda _{\mathcal{S}}$ = 40. First, the number of labeled data is set to 2,500. The result is shown in the fourth column of Table \ref{tab1}. 
\begin{wraptable}{tp}{7.1cm}
\caption{Classification accuracy on CIFAR-100 under different backbones.}
\vspace{-1em}
\label{tab2}
\renewcommand\arraystretch{1.05}
\centering
\setlength{\tabcolsep}{0.1mm}
\begin{tabular}{c|c|c}
\specialrule{0em}{4pt}{0pt}
\hline
\multicolumn{3}{c}{CIFAR-100}\\
\hline
Methods &10000 labels&Backbone\\
\hline
\hline
MixMatch \cite{berthelot2019mixmatch} &64.01\%&WRN28-2\\
MixMatch Enhanced \cite{hu2021simple} &67.12\%&WRN28-2\\
SimPLE \cite{hu2021simple}&70.82\%&WRN28-2\\
\textbf{Proposed ADT-SSL}&\textbf{74.14\%}&\textbf{WRN28-2}\\
\hline
$\Pi$-Model \cite{rasmus2015semi}&62.12\%&WRN28-8\\
Pseudo-Labeling \cite{lee2013pseudo}&63.79\%&WRN28-8\\
Mean Teacher \cite{tarvainen2017mean}&64.17\%&WRN28-8\\
MixMatch \cite{berthelot2019mixmatch}&71.69\%&WRN28-8\\
ReMixMatch \cite{berthelot2019remixmatch}&76.97\%&WRN28-8\\
UDA \cite{xie2020unsupervised}&75.50\%&WRN28-8\\
FixMatch \cite{sohn2020fixmatch}&77.40\%&WRN28-8\\
SimPLE \cite{hu2021simple}&78.11\%&WRN28-8\\
\textbf{Proposed ADT-SSL}&\textbf{79.32\%}&\textbf{WRN28-8}\\
\hline
Fully-Supervised&80.93\%&WRN28-8\\
\hline
\end{tabular}
\end{wraptable}
Clearly, our ADT-SSL achieves the best performance among all the compared SSL methods and exceeds its best competitor by 1.14\%. Since CIFAR-100 has 100 classes and is a relatively complex dataset, all the compared methods perform much worse than the fully-supervised method. Also note that the fully-supervised method on CIFAR-100 is much worse than on CIFAR-10. As we use more labeled data, the SSL methods should show a clear enhancement. The result on 10,000 labeled data is shown in Table \ref{tab2}. We can see from Table \ref{tab2} that testing on both WRN28-2 and WRN28-8, our method surpasses all the baselines by a large margin. By comparing Table \ref{tab2} with the fourth column of Table \ref{tab1}, we clearly see that all the compared SSL methods obtain a significant improvement when using more labeled data. Even for the complex CIFAR-100 dataset, our ADT-SSL also approaches to the fully-supervised method on the Wide ResNet 28-8 backbone. Compared to the leading SimPLE \cite{hu2021simple}, our ADT-SSL achieves over 1\% improvement. It is worth emphasizing that ADT-SSL surpasses its best competitor by 3.32\% on the Wide ResNet 28-2 backbone. This result demonstrates that ADT-SSL can deal with more difficult classification tasks and perform well on the lightweight backbone. This is due to the fact that the proposed class-adaptive threshold can balance the learning states of the model for each class. When the number of labeled data increases, ADT-SSL can better adjust the model training via $\mathcal{L} _{\mathcal{X}}$. More importantly, because the class-adaptive threshold is extracted during the model training, the increase of labeled data means that the class-adaptive threshold can better reflect the learning state of each class.

\noindent
\textbf{Results on SVHN:} For SVHN, we empirically set $\lambda _{\mathcal{U}1}=4$, $\lambda _{\mathcal{U}2}=50$, and $\lambda _{\mathcal{S}}=16$. The result is shown in Table \ref{tab1}. We can see from the last two columns of Table \ref{tab1} that ADT-SSL does not perform best. But it is interesting that some SSL methods, such as UDA \cite{xie2020unsupervised}, ReMixMatch \cite{berthelot2019remixmatch}, FixMatch \cite{sohn2020fixmatch}, SimPLE \cite{hu2021simple}, and our ADT-SSL, outperform the fully-supervised method. The reasons might be as follows: 1) SVHN consists of regular digits and thus is a relatively simple dataset. As a result, both the fully-supervised and semi-supervised methods can gain satisfactory performance on this dataset. Fortunately, most SSL methods adopt the data augmentation technology, thereby increasing the number of data and enlarging the diversity of data significantly. This results in a performance improvement of these SSL methods against the fully-supervised method. 2) All the existing SSL methods use a very high fixed threshold (e.g., 0.95) for pseudo-labeling the unlabeled data. Most models can output a high confidence for the unlabeled data due to easy recognition of regular digits. Therefore, this fixed threshold can pick out the most valuable unlabeled data. On the contrary, ADT-SSL extracts a lowest threshold of each class from the labeled data which can be correctly classified. Using this extracted threshold to pseudo label those unlabeled data could draw a certain amount of noisy labels into the model training, thus leading to performance degradation.
\begin{table*}[tp]
\renewcommand\arraystretch{1.1}
\centering
\setlength{\tabcolsep}{2.2mm}
\caption{Ablation study on CIFAR-100 with the Wide ResNet 28-2 backbone.}
\label{tab3}
\vspace{-0.5em}
\begin{tabular}{l|l|c|c|c|c|c}
\hline
Ablation&Class-Adaptive Threshold&$\lambda _{\mathcal{S}}$&$\tau $&$\mathcal{T} _s$&$K$&10000 labels\\
\hline
ADT-SSL&Adopt&40&0.95&0.9&2&74.14\%\\
ADT-SSL&Adopt&40&0.95&0.9&7&\textbf{75.43}\%\\
w/o $\mathcal{T} _c$&Without&40&0.95&0.9&2&73.58\%\\
w/o $\mathcal{T} _c$&Without&40&0.95&0.9&7&74.28\%\\
w/o Similar Loss&Adopt&0&0.95&0.9&2&73.63\%\\
w/o Similar Loss&Adopt&0&0.95&0.9&7&73.78\%\\
w/o $\mathcal{T} _c$, w/o Similar Loss&Without&0&0.95&0.9&2&70.99\%\\
\hline
$\tau$=0.75, $\mathcal{T} _s$=0.9&Adopt&40&0.75&0.9&2&74.48\%\\
$\tau$=0.95, $\mathcal{T} _s$=0.7&Adopt&40&0.95&0.7&2&74.51\%\\
$\tau$=0.75, $\mathcal{T} _s$=0.7&Adopt&40&0.75&0.7&2&74.41\%\\
\hline
\end{tabular}
\end{table*}
\subsection{Ablation Study}
In this part, we conduct an ablation study to analyze the contribution of each component of the proposed ADT-SSL. We use Wide ResNet 28-2 as the backbone and test on CIFAR-100 with 10,000 labeled data. The result is shown in Table \ref{tab3}. The performance of the proposed ADT-SSL is greatly increased as the number of augmentation forms increases. It shows that the proposed ADT-SSL can adapt to multiple forms of augmentations and learn more useful information from them. When we abandon either class-adaptive threshold or similar loss, there is a drop in classification accuracy. Especially when we simultaneously abandon both class-adaptive threshold and similar loss, the number of the unlabeled data participated in the training process will be dramatically reduced, resulting in a significant performance degradation. This demonstrates that the proposed class-adaptive threshold and similar loss can both contribute to the SSL method substantially. In addition, we test the performance of ADT-SSL on different values $\tau$ and $\mathcal{T} _s$. We can see from the last three rows of Table \ref{tab3}, ADT-SSL remains a stable classification accuracy. This indicates ADT-SSL is robust to changes in the threshold values. This is because adjusting a single threshold of ADT-SSL will not affect the total number of unlabeled data participating in training. Therefore, there is no change in the overall loss.

\section{CONCLUSIONS}
In this paper, we have presented an Adaptive Dual-Threshold method for Semi-Supervised Learning (ADT-SSL). The major contributions of the paper can be summarized as follows: 1) We find that a large amount of unlabeled data whose predictions are below the fixed threshold have not been considered in existing SSL methods while these data are often of hard samples, enabling to benefit the model training. 2) A class-adaptive threshold method is proposed for the first time, which makes a large amount of unlabeled data be added to the model training, thus significantly boosting the performance of the SSL methods. 3) A novel similar loss is designed for learning from the highly similar unlabeled data, which makes the prediction of the model more consistent. 4) We propose to assign the four different loss functions to four different types of data respectively, namely supervised loss for labeled data, one unsupervised loss for unlabeled data above the fixed threshold, the other unsupervised loss for unlabeled data below the fixed threshold but above the adaptive threshold, and similar loss for highly similar unlabeled data. 5) Both extensive experimental results and ablation study have demonstrated the proposed ADT-SST is on par with the state-of-the-art methods. Our findings suggest that massive valuable information can be mined from indefinitely extended unlabeled data, which provides a promising direction to boost the performance of SSL.

\bibliographystyle{IEEEtran}  
\bibliography{IEEEexample}  

\begin{thebibliography}{10}
\providecommand{\url}[1]{#1}
\csname url@samestyle\endcsname
\providecommand{\newblock}{\relax}
\providecommand{\bibinfo}[2]{#2}
\providecommand{\BIBentrySTDinterwordspacing}{\spaceskip=0pt\relax}
\providecommand{\BIBentryALTinterwordstretchfactor}{4}
\providecommand{\BIBentryALTinterwordspacing}{\spaceskip=\fontdimen2\font plus
\BIBentryALTinterwordstretchfactor\fontdimen3\font minus
  \fontdimen4\font\relax}
\providecommand{\BIBforeignlanguage}[2]{{%
\expandafter\ifx\csname l@#1\endcsname\relax
\typeout{** WARNING: IEEEtran.bst: No hyphenation pattern has been}%
\typeout{** loaded for the language `#1'. Using the pattern for}%
\typeout{** the default language instead.}%
\else
\language=\csname l@#1\endcsname
\fi
#2}}
\providecommand{\BIBdecl}{\relax}
\BIBdecl

\bibitem{laine2016temporal}
S.~Laine and T.~Aila, ``Temporal ensembling for semi-supervised learning,''
  \emph{arXiv preprint arXiv:1610.02242}, 2016.

\bibitem{sajjadi2016regularization}
M.~Sajjadi, M.~Javanmardi, and T.~Tasdizen, ``Regularization with stochastic
  transformations and perturbations for deep semi-supervised learning,''
  \emph{Advances in Neural Information Processing Systems}, vol.~29, pp.
  1171--1179, 2016.

\bibitem{sohn2020fixmatch}
K.~Sohn, D.~Berthelot, N.~Carlini, Z.~Zhang, H.~Zhang, C.~A. Raffel, E.~D.
  Cubuk, A.~Kurakin, and C.-L. Li, ``Fixmatch: Simplifying semi-supervised
  learning with consistency and confidence,'' \emph{Advances in Neural
  Information Processing Systems}, vol.~33, pp. 596--608, 2020.

\bibitem{berthelot2019mixmatch}
D.~Berthelot, N.~Carlini, I.~Goodfellow, N.~Papernot, A.~Oliver, and C.~A.
  Raffel, ``Mixmatch: A holistic approach to semi-supervised learning,''
  \emph{Advances in Neural Information Processing Systems}, vol.~32, pp.
  5049--5059, 2019.

\bibitem{berthelot2019remixmatch}
D.~Berthelot, N.~Carlini, E.~D. Cubuk, A.~Kurakin, K.~Sohn, H.~Zhang, and
  C.~Raffel, ``Remixmatch: Semi-supervised learning with distribution alignment
  and augmentation anchoring,'' \emph{arXiv preprint arXiv:1911.09785}, 2019.

\bibitem{miyato2018virtual}
T.~Miyato, S.-I. Maeda, M.~Koyama, and S.~Ishii, ``Virtual adversarial
  training: A regularization method for supervised and semi-supervised
  learning,'' \emph{IEEE Transactions on Pattern Analysis and Machine
  Intelligence}, vol.~41, no.~8, pp. 1979--1993, 2018.

\bibitem{tarvainen2017mean}
A.~Tarvainen and H.~Valpola, ``Mean teachers are better role models:
  Weight-averaged consistency targets improve semi-supervised deep learning
  results,'' \emph{Advances in Neural Information Processing Systems}, vol.~30,
  2017.

\bibitem{xie2020unsupervised}
Q.~Xie, Z.~Dai, E.~Hovy, T.~Luong, and Q.~Le, ``Unsupervised data augmentation
  for consistency training,'' \emph{Advances in Neural Information Processing
  Systems}, vol.~33, pp. 6256--6268, 2020.

\bibitem{chen2020simple}
T.~Chen, S.~Kornblith, M.~Norouzi, and G.~Hinton, ``A simple framework for
  contrastive learning of visual representations,'' \emph{International
  Conference on Machine Learning}, vol. 119, pp. 1597--1607, 2020.

\bibitem{zhang2020wcp}
L.~Zhang and G.-J. Qi, ``Wcp: Worst-case perturbations for semi-supervised deep
  learning,'' \emph{IEEE/CVF Conference on Computer Vision and Pattern
  Recognition}, pp. 3912--3921, 2020.

\bibitem{hu2021simple}
Z.~Hu, Z.~Yang, X.~Hu, and R.~Nevatia, ``Simple: Similar pseudo label
  exploitation for semi-supervised classification,'' \emph{IEEE/CVF Conference
  on Computer Vision and Pattern Recognition}, pp. 15\,099--15\,108, 2021.

\bibitem{zhang2021flexmatch}
B.~Zhang, Y.~Wang, W.~Hou, H.~Wu, J.~Wang, M.~Okumura, and T.~Shinozaki,
  ``Flexmatch: Boosting semi-supervised learning with curriculum pseudo
  labeling,'' \emph{Advances in Neural Information Processing Systems},
  vol.~34, pp. 18\,408--18\,419, 2021.

\bibitem{cubuk2019autoaugment}
E.~D. Cubuk, B.~Zoph, D.~Mane, V.~Vasudevan, and Q.~V. Le, ``Autoaugment:
  Learning augmentation strategies from data,'' \emph{IEEE/CVF Conference on
  Computer Vision and Pattern Recognition}, pp. 113--123, 2019.

\bibitem{cubuk2020randaugment}
E.~D. Cubuk, B.~Zoph, J.~Shlens, and Q.~V. Le, ``Randaugment: Practical
  automated data augmentation with a reduced search space,'' \emph{IEEE/CVF
  Conference on Computer Vision and Pattern Recognition Workshops}, pp.
  702--703, 2020.

\bibitem{chapelle2009semi}
O.~Chapelle, B.~Scholkopf, and A.~Zien, ``Semi-supervised learning (chapelle,
  o. et al., eds.; 2006)[book reviews],'' \emph{IEEE Transactions on Neural
  Networks}, vol.~20, no.~3, pp. 542--542, 2009.

\bibitem{lee2013pseudo}
D.-H. Lee \emph{et~al.}, ``Pseudo-label: The simple and efficient
  semi-supervised learning method for deep neural networks,''
  \emph{International Conference on Machine Learning Workshop}, vol.~3, no.~2,
  pp. 896--896, 2013.

\bibitem{douze2018low}
M.~Douze, A.~Szlam, B.~Hariharan, and H.~J{\'e}gou, ``Low-shot learning with
  large-scale diffusion,'' \emph{IEEE/CVF Conference on Computer Vision and
  Pattern Recognition}, pp. 3349--3358, 2018.

\bibitem{iscen2019label}
A.~Iscen, G.~Tolias, Y.~Avrithis, and O.~Chum, ``Label propagation for deep
  semi-supervised learning,'' \emph{IEEE/CVF Conference on Computer Vision and
  Pattern Recognition}, pp. 5070--5079, 2019.

\bibitem{bhattacharyya1946measure}
A.~Bhattacharyya, ``On a measure of divergence between two multinomial
  populations,'' \emph{Sankhy{\=a}: The Indian Journal of Statistics
  (1933-1960)}, vol.~7, no.~4, pp. 401--406, 1946.

\bibitem{krizhevsky2009learning}
A.~Krizhevsky, G.~Hinton \emph{et~al.}, ``Learning multiple layers of features
  from tiny images,''
  \url{http://citeseerx.ist.psu.edu/viewdoc/download?doi=10.1.1.222.9220&rep=rep1&type=pdf},
  2009.

\bibitem{netzer2011reading}
Y.~Netzer, T.~Wang, A.~Coates, A.~Bissacco, B.~Wu, and A.~Y. Ng, ``Reading
  digits in natural images with unsupervised feature learning,''
  \url{http://ufldl.stanford.edu/housenumbers/nips2011_housenumbers.pdf}, 2011.

\bibitem{rasmus2015semi}
A.~Rasmus, M.~Berglund, M.~Honkala, H.~Valpola, and T.~Raiko, ``Semi-supervised
  learning with ladder networks,'' \emph{Advances in Neural Information
  Processing Systems}, vol.~28, pp. 596--608, 2015.

\bibitem{zagoruyko2016wide}
S.~Zagoruyko and N.~Komodakis, ``Wide residual networks,'' \emph{arXiv preprint
  arXiv:1605.07146}, 2016.

\bibitem{loshchilov2016sgdr}
I.~Loshchilov and F.~Hutter, ``Sgdr: Stochastic gradient descent with warm
  restarts,'' \emph{arXiv preprint arXiv:1608.03983}, 2016.

\end{thebibliography}

\end{document}